# Early-Phase Performance-Driven Design using Generative Models

Spyridon Ampanavos[1,2]* and Ali Malkawi[1,2]

[1]Harvard Graduate School of Design, Cambridge MA 02138, USA
[2]Harvard Center for Green Buildings and Cities, Cambridge MA 02138, USA

`sampanavos@gsd.harvard.edu, amalkawi@gsd.harvard.edu`

**Abstract.** Current performance-driven building design methods are not widely adopted outside the research field for several reasons that make them difficult to integrate into a typical design process. In the early design phase, in particular, the time intensity and the cognitive load associated with optimization and form parametrization are incompatible with design exploration, which requires quick iteration. This research introduces a novel method for performance-driven geometry generation that can afford interaction directly in the 3d modeling environment, eliminating the need for explicit parametrization, and is multiple orders faster than the equivalent form optimization. The method uses Machine Learning techniques to train a generative model offline. The generative model learns a distribution of optimal performing geometries and their simulation contexts based on a dataset that addresses the performance(s) of interest. By navigating the generative model's latent space, geometries with the desired characteristics can be quickly generated. A case study is presented, demonstrating the generation of a synthetic dataset and the use of a Variational Autoencoder (VAE) as a generative model for geometries with optimal solar gain. The results show that the VAE-generated geometries perform on average at least as well as the optimized ones, suggesting that the introduced method shows a feasible path towards more intuitive and interactive early-phase performance-driven design assistance.

## 1 Introduction

During the design process, an architect strives to reconcile several qualitative and quantitative objectives. Performance-driven design aims to assist in meeting the quantifiable objectives related to a building's performance, most commonly through the use of optimization. To maximize its impact, the performance-driven design methodology needs to be applied from the early design phase[1]. Contrary to its original purpose as a precise problem-solving tool, optimization is increasingly gaining traction as an exploratory tool in the early design phase[4–10]. However, outside of the research field, the use of optimization in the early design phase has been limited for reasons that relate to: i) time intensity, ii) interpretability, iii) inherent limitations of the required parametric models, and iv) the elusive nature of performance goals in architectural design.

---

[1] Paulson and MacLeamy have both elaborated on the impact of changes along the different phases of design [1, 2]. Similarly, Morbitzer argues that simulation should be used throughout the design process [3].

*Time Intensity.* One major limitation for applying optimization in architecture is the time intensity of the processes involved [11, 6, 12–14]. Environmental or structural simulations can be computationally expensive. Combined with an optimization process that employs a stochastic search method, such as evolutionary algorithms, the calculation time increases by multiple orders of magnitude. In the early design phase, where it is essential to consider multiple design alternatives quickly, the slow speed of optimization disrupts the exploratory process.

*Interpretability.* When it comes to interpreting multi-objective optimization results, architects can have difficulties in understanding the solution space [15, 16]. Optimization returns a set of high-performing solutions with their corresponding performances; however, the connection between design parameters and performance tradeoffs is not always apparent [16], offering little intuition to the designer.

*Limitations of Parametric Models.* Parametric models are widely adopted in architecture; however, their applicability in the early design phase has been questioned [4, 17–19]. Davis offers an extensive analysis of how parametric models have certain limits on the changes they can afford before breaking [20]. To accommodate a major change, such as those that often happen during the conceptual phase, the parametric model would need to be replaced by a new one [9, 21, 22]. However, optimization operates on a pre-determined parametric model, and as a result, it conflicts with the nature of the conceptual design stage.

*Nature of Performance Goals in Design.* Carrara et al. describes the design process as consisting of three operations [23]: i) definition of the desired set of performance criteria, ii) production of design solutions, and iii) evaluation of expected performance. However, they stress that these operations relate in a non-linear way and coevolve during the design process. Others have also discussed the co-evolution of the problem definition and solution in the design process [11, 12, 16, 18, 24, 25]. Therefore, it is expected that the performance goals, parameters, and constraints will be redefined multiple times during the design process.

Consequently, as long as optimization requires from the designer a high investment in terms of time and cognitive effort to create the parametric abstractions and interpret the results, it cannot seamlessly integrate into the early design phase.

This research suggests an alternative method of providing early-phase performance-driven design assistance for optimally performing geometries in real-time and without the need for parametrization. The method makes use of Machine Learning (ML) generative models. It relies on the navigation of a latent space where the results of a series of optimization processes have been encoded in advance.

This paper describes the suggested method and presents a case study where a Variational Auto-Encoder (VAE) is introduced for the generation of geometries with optimal solar gain properties and pre-determined size. The results show that the VAE was able to generate geometries with optimal or close-to-optimal performance for most simulation contexts from the test set, in a fraction of the corresponding optimization time.

This work makes the following contributions to the area of computational design. A novel method for real-time early-phase performance-driven design assistance is introduced, which does not require parametric models. The use of generative models is suggested for the novel task of generating geometries with optimal performance properties. Empirical evidence is provided, suggesting that a VAE can be used as a generative model for optimally performing geometries.

## 2    Related Work

### 2.1    Performance-Driven Design

**Simulation.** Simulations form the basis of performance-driven design. However, a single or a limited number of simulations is not enough to guide design improvements. Systematic simulations [26] attempted to address this subject, but the complex relationship between performance and parameters related to form, together with the time intensity of the calculations, make this an impractical solution. Some template-based tools attempted to give a solution by enabling quick evaluation of alternatives [3, 27, 28], however, they imposed severe restrictions to the range of supported forms and thus were not adopted by the architectural community. Finally, real-time simulations were found to be helpful during the performance-driven form-finding process [29], however, in cases with large design spaces and multiple performance criteria, further guidance is necessary [30].

**Sensitivity Analysis.** Sensitivity analysis methods can be used to guide exploration based on a single parametric model [31, 32] or to evaluate multiple alternative parametric models [33–35]. However, it has been argued that they do not provide adequate information to lead to optimally performing solutions [13, 36].

**Optimization.** Optimization processes identify the parameters of a model that result in optimally performing solutions. They have successfully solved engineering or building science problems [5, 6, 37]. However, despite extensive research on optimization for performance-driven design, such methods have not been widely adopted in the architecture practice [5, 6].

**Form Exploration.** In order to reconcile the engineering nature of optimization with the more exploratory role that designers tend to give to it [4], some research has suggested interactive optimization [8, 9] for integrating performance with designer preferences. Other work has focused on simulation speed, interactivity, and results visualization [16] through the use of surrogate modeling. Last, some recent work has suggested eliminating the parametric modeling overhead by deploying automatic parametrization and data analysis [17].

## 2.2 Generative Models

**Definition.** A generative model is a type of ML model that can learn an estimate of a distribution by observing a set of examples from that distribution, i.e., a training set [38]. Once fully trained, sampling a generative model approximates sampling from the original data distribution. For example, a generative model trained on a dataset of faces will generate new faces when sampled[2].

**Latent Space.** Some generative models work by learning a mapping of the original data to a lower-dimensional space, called the latent space. For example, the Variational Auto-Encoder [41] (VAE) explicitly learns an encoder and a decoder function that maps the original data to and from a latent space. Naturally, similar data points will be located close in the latent space. This characteristic allows for smooth interpolation of data samples by traversing the latent space or even for the composition of new data with specified properties through latent space vector arithmetic, as demonstrated by Wu et al. in the domain of three-dimensional objects [42].

**Applications.** In the field of architecture, several attempts have been made to use generative models in the creative phase [43–46]. Most such works used Generative Adversarial Networks [47] (GANs), motivated by some impressive results in the field of computer vision [39, 47–49]. However, the subject of performance has not been previously addressed directly in research related to generative models.

## 3 Approach

Current practices and previous research reveal a lack of support for performance-driven design in the early form-finding process. Almost all related research approaches performance-driven design through the scope of parametric modeling, which imposes severe restrictions of time-intensity, cognitive load, and premature commitment to specific graph topologies[19, 22] to the creative process.

This research suggests that optimal form-finding can be achieved by navigating the latent space of a generative model. A generative model that addresses a specific set of performance metrics can be trained on a dataset where each data point represents both the problem definition and an optimal solution to the problem. When the trained model is sampled, it will generate a new problem definition and an optimal solution following the learned data distribution. In order to generate an optimal solution to a specific problem definition, a sample can be retrieved from the model, constrained by the problem definition of interest. In practice, this can be achieved through search or navigation of the model's latent space. In addition to generating optimal geometric forms from scratch, the same generative model can also be used to suggest optimally performing alternatives that are as close as possible to user-generated forms. For this task, the user-generated geometry becomes part of the constraints that drive the latent space search.

---

[2]   See for example the Progressive GAN model [39] trained on the CelebA dataset [40].

The proposed method addresses the current limitation of time intensity associated with performance-driven design. In addition, the ability of ML methods to deal with high dimensional data is used to work directly with geometries from the modeling environment, eliminating the need for geometry parametrization. Last, the method opens up the potential for intuitive, real-time interactivity in a user-guided search for optimal geometries.

The dataset required to train such a model needs to include a diverse set of optimally performing geometries for a wide range of problem definitions. Such datasets do not exist at the moment and are impossible to collect from the real world, so synthetic datasets with the desired characteristics should be created using existing optimization methods. Since a specific model only addresses a pre-determined set of performance metrics, the term "problem definition" refers to the simulation context that drives the optimization process.

Next, a case study is presented, where the performance of interest is related to the solar gain and the size of the building. The case study allows a detailed development and evaluation of the suggested techniques.

## 4 Case Study

### 4.1 Problem Scope

In a typical scenario for the design of a new building, the architect would have information including the location, the plot shape and size, the surrounding buildings, and the program of the building. In performance-driven design, maximizing the performance of interest is of primary concern. Then, in the early design phase, where the focus is on form finding, the problem would be expressed as finding a geometry for the building that maximizes the desired performance, given the simulation context (Figure 1).

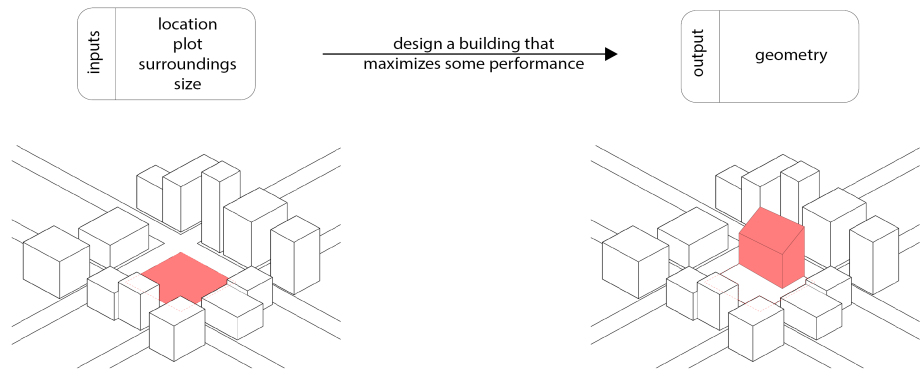

**Fig. 1.** Optimization: context and expectation for the design of a new building.

In this case study the goal was to generate a geometric form for a building that maximizes the average solar radiation gain during wintertime while keeping the size as close as possible to a predefined target. In more detail, the solar gain objective was defined as the average of the received radiation per area unit of all the mesh faces of a generated geometry. The size was represented by the geometry volume. The location of the building was Boston. Wintertime for the environmental simulation purposes of this case study was defined as any time when the temperature is below 12 C. The plot shape was a square with a side of 10m. In addition, a maximum height of 10m for any building was set.

In this problem, the term boundary condition refers to the configuration of the surrounding buildings, as this was the only part of the solar simulation's boundary condition that varied. The range of the boundary condition was up to one obstructing building on each of the east, south, and west sides of the plot, and up to three total obstructing buildings (Figure 2). With all obstructions having the same width and height, a total of 342 unique boundary conditions were used.

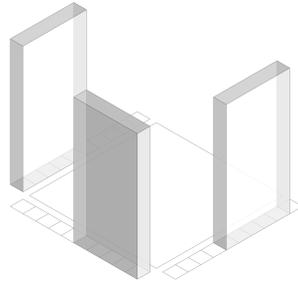

**Fig. 2.** Range of variable boundary conditions of the simulation. Each of the three parallelepipeds can move on the outlined locations or be omitted.

### 4.2 Data Generation

**Geometry Representation.** An optimization algorithm was used during the data generation phase together with a parametric model that generated the geometry. In architectural design, it is common for a parametric model to be created using high-level concepts, such as box or tower, and their transformations, such as scale or twist angle. However, a more neutral and low-level geometry representation is more suitable when no conceptual decisions are assumed. Therefore, a heightmap was used as the geometry generation model. Each parameter of the model controls the height of a point on a two-dimensional grid. This representation provides a simple and intuitive way to describe geometries, with a fair amount of flexibility. One limitation is that it cannot describe certain three-dimensional forms. For example, a height map cannot encode information about cantilevers.

**Optimization.** *Objective.* As described in the problem definition, the performance goal was to maximize the solar gain during the winter months. At the same time, the total volume was constrained to remain as close as possible to a predefined target (v=100m$^3$). The volume constraint was used as a proxy for the architectural program, which would prescribe the total surface area in a real-life scenario. In practice, the volume constraint was transformed to a minimization objective, calculated as the squared difference of the target volume from the current geometry volume.

When optimizing a problem with multiple objectives, there are two major approaches. The first is to use scalarization with a single objective algorithm; the second is to use a multi-objective algorithm. This case study used scalarization, combining the two objectives into a single, minimization objective as described in equation (1), where a geometry is a mesh instance, J is the minimization objective, AvgRadiation(geometry) evaluates the average of the wintertime incident radiation on all the faces of the geometry mesh, Vol_target is the desired volume, and Vol(geometry) calculates the volume of a geometry.

$$J(\text{geometry}) = -\text{AvgRadiation(geometry)} + (\text{Vol\_target} - \text{Vol(geometry)})^2 * 10^{-3} \quad (1)$$

*Optimal Solutions Selection.* When solving a problem with multiple objectives, the Pareto front, i.e., the set of non-dominated solutions, has been commonly used to identify the best-performing solutions [15, 50–55]. Therefore, the individual objectives on each step of the optimization were recorded, and after the optimization was complete, the Pareto front was calculated, as suggested in relevant work [56]. For each optimization problem, i.e., for each of the 342 boundary conditions, a total of 10 optimal results were selected to form a dataset.

*Implementation.* The solar radiation calculation was performed using the open-source plugin Ladybug [57], inside the visual programming platform Grasshopper3d in McNeel's Rhinoceros 3d modeling software. A communication module for Grasshopper was developed using web sockets, connecting the parametric model and the solar simulations to an external optimization algorithm. The optimization algorithm was a customized implementation of Simulated Annealing (SA). The whole workflow was controlled by a command-line program that called the optimization algorithm and obtained the performance results from Grasshopper.

The 342 optimizations were run on a desktop computer for a fixed number of optimization steps (n=3000). Each optimization required an average of approximately 20 minutes to complete. After selecting the ten best solutions for each boundary condition, a dataset of 3420 pairs of boundary condition – optimal geometry was created.

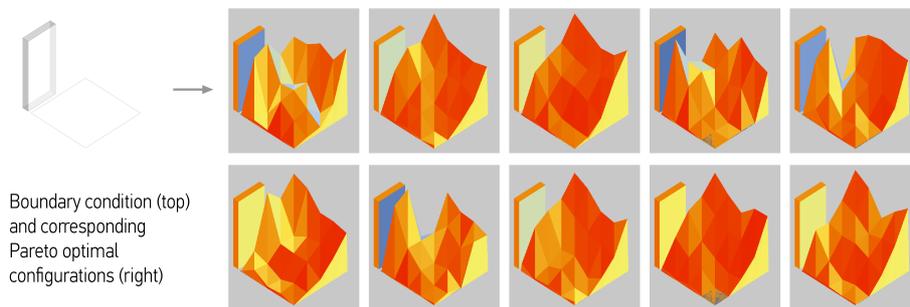

Boundary condition (top) and corresponding Pareto optimal configurations (right)

**Fig. 3.** Example of a boundary condition and the selected Pareto optimal geometries.

### 4.3 Learning

**Generative Models.** Two popular types of generative models in ML are the Generative Adversarial Networks (GAN) and the Variational Auto-encoders (VAE). While either type could be used with the suggested method, this case study focuses on the use of a VAE because of the simpler setup and its natural ability to model a latent space. This last feature is essential since it is the navigation of this latent space that enables sampling optimal geometries for specific boundary conditions.

**Data Format.** In order to use the generated data with an ML model they first had to be converted to vectors. Since the boundary conditions in this problem are geometric, both boundary conditions and optimal geometries were incorporated into a single geometric representation. In ML, there are three primary ways to describe geometric data [58]: i) image-based (single or multi-view), ii) voxel-based, and iii) point clouds. Image-based methods are currently the most robust and well-developed methods and compared to the original parametric description of the geometries, they allow better modeling of the spatial relationships between the individual parameters of the vector representation through the use of convolutions in the learning model.

Since all geometries in the dataset were created using a heightmap, a single depth map from a top view was used to describe the data (Figure 4). Multiple different image resolutions were considered for the depth map before a resolution of 16X16 pixels was selected based on initial results when using the VAE.

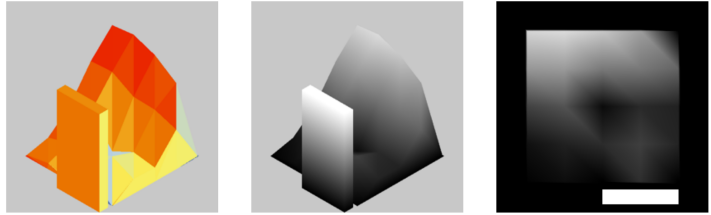

**Fig. 4.** From left to right: i) optimized geometry with solar radiation colors – SE Isometric, ii) optimized geometry with depth map – SE Isometric, iii) top view projection of (ii) – final format of data.

**Using the Variational Auto-encoder (VAE).** A VAE is a probabilistic model that learns an encoder function E(x) and a decoder function D(z), mapping from the original data to a lower-dimensional latent space and back to the original data by training on a reconstruction task. The objective is defined as the reconstruction loss with a regularizer. The reconstruction loss encourages the decoder to learn to reconstruct the data. The regularizer is the Kullback-Leibler divergence of the approximate posterior (i.e., the encoder's distribution) from the prior (commonly chosen as a Gaussian distribution) [41]. Equations (2) and (3) provide a simplified description of the VAE in terms of the encoder and decoder functions, where x is an input vector, z is the mapping of x in the latent space, $g_\varphi$ is a reparametrization function[3], and y is the reconstruction of x.

$$z = E(x), z' = g_\varphi(z), y = D(z') \qquad (2)$$

$$y = D(g_\varphi(E(x))) \qquad (3)$$

After a VAE has been trained, the decoder function can be isolated and used as a generative model that samples the latent space and generates data instances from the original distribution. In this work, the decoder is used to generate images (depth maps) of boundary conditions and corresponding optimal geometries.

The retrieval of data instances for a specific boundary condition was achieved as follows. First, a loss function Lb was defined as the distance of a generated instance's boundary condition from the desired boundary condition, described in (4). Next, the boundary condition – i.e., the surrounding buildings abstracted to simple parallelepipeds – was translated to a depth map, following the same data format on which the VAE was trained, but without any corresponding optimal geometry. The desired geometries were found by solving the optimization problem (5) of finding the sample z in the latent space, for which the decoder produces a depth map that minimizes the boundary condition loss $L_b$.

$$L_b(\text{target\_boundary}, y) = \text{Distance}(\text{target\_boundary}, \text{Boundary\_Condition}(y)) \qquad (4)$$

$$J(z) = L_b(\text{target\_boundary}, D(z)) \qquad (5)$$

Since the decoder – and consequently the loss $L_b$ – is a differentiable function, problem (5) can be solved using gradient descent. Vector z is initialized as a random sample of the latent space. The loss $L_b$ is calculated, and its gradient is backpropagated to the decoder's input, resulting in an update of z. Several updates are performed, until convergence. Using gradient descent in this process is of particular importance because it enables high-speed retrieval of the appropriate latent space vector, in contrast to alternative search methods such as stochastic sampling.

**Model Architecture**. *Training*. The VAE was implemented as a convolutional neural net (Figure 5). The encoder consists of two convolutional layers followed by a fully connected layer with output size 32. This output corresponds to the mean and standard deviation of a normal distribution of dimension 16, so the latent space is 16-dimensional. The decoder – or generative model – follows a mirrored structure of the encoder. The input-output of the VAE is a 16X16 grayscale image. The reconstruction loss was defined as the L2 distance (squared difference) of the input-output images.

---

[3]  The reparametrization is an essential component of the VAE, but only mentioned here for reasons of completeness. For details we direct the interested reader to [41].

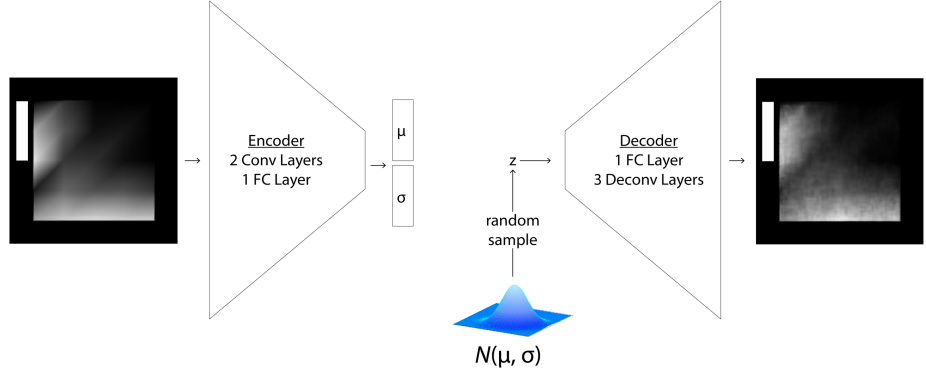

**Fig. 5.** VAE architecture used for training.

*Inference.* For the process of finding appropriate boundary condition – optimal geometry instances for a specific boundary condition, a boundary condition loss function was defined. This function calculates how close a generated image's boundary condition is to the desired boundary condition. The generated image is first masked to only leave the boundary condition visible. Then, the masked image is compared to the desired boundary condition image (Figure 6). The distance of the two images was calculated using the sigmoid cross-entropy.

**Training.** The total dataset was split into two parts, a training set containing 90% of the data (3,080 data points that belong to 308 boundary conditions) and a test set containing the rest 10% of the data (340 data points that belong to 34 unique boundary conditions). The split was done through random selection and care to place all ten data points from the same boundary condition in the same group, ensuring that the boundary conditions found in the test dataset have not been encountered during the training. The VAE model was implemented using the Python library TensorFlow [59] and trained for 1000 epochs, using the Adam optimizer [60] and batch size 32. The loss function was implemented as the single sample Monte Carlo estimate of the expectation [61], where the reconstruction loss is the squared difference of the input-output images. Only minor improvements in the loss were gained between 200 and 1000 epochs. At 1000 epochs, a validation loss of 9.3 was achieved.

**Inference**. Inferred geometries were generated for the 34 unique boundary conditions of the test set using gradient descent. Due to the random initialization of the process and the non-convex shape of the latent space, different geometries can be obtained for the same boundary condition through repeated optimizations. For each of the boundary conditions, 100 geometries were generated. The optimization algorithm Adam was used with a learning rate of 0.02 for 400 iterations. Convergence was typically observed in less than 200 iterations.

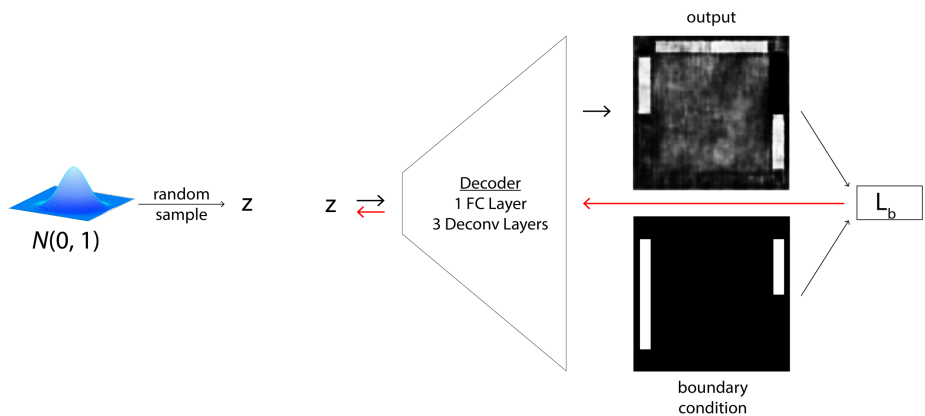

**Fig. 6.** Inference using the decoder part of the VAE.

## 5 Results

The method introduced for predicting optimal geometries for specific boundary conditions relies on the two underlying processes: i) a high-quality mapping of the data to a latent space, and ii) the successful navigation of this low-dimensional space.

The first process requires a mapping function that can encode all the critical information, as well as a well-structured latent space that enables the generation of new data through successful interpolations. Both processes are evaluated by assessing the quality of reconstructions that the VAE produces for the test set. If the VAE was used for a visual task, the reconstruction quality would refer to the similarity of the input and output images. However, since the overall goal of this problem relates to building performance, the evaluation was performed with respect to the specific performance goals of solar radiation and volume compliance, as they have been detailed in Section 4.2. The geometries derived from the VAE-reconstructed depth maps are expected to perform close to the optimization-derived geometries.

Similarly, the navigation of the latent space is evaluated based on the performance of the geometries generated, constrained by the boundary conditions in the test set, using the process described in Section 4.3.

Potential inaccuracies in the actual performance metrics of the dataset may have been introduced during the resampling process, when meshes based on a 5X5 height-map were encoded to 16X16 depth map images. To avoid this issue when comparing SA-optimized with VAE-generated geometries, the reported performance of both the test set ground truth and the test set inferences was calculated following a common process based on the 16X16 depth map encodings of the geometry.

### 5.1 Reconstruction Performance

The reconstruction results for all 34 boundary conditions of the test set were coded into three categories after careful observation of the per-boundary condition scatterplots and a comparison of the mean performances. The results for 5 boundary conditions were coded as type a: performance very close to the test set, 27 were coded as type b: performance on one axis similar to the test set and the other axis better than the test set, and 2 were coded as type c: performance better than the test set. Diagrams a, b, and c of Figure 7 show a representative sample from each type. While some individual geometries with bad overall performance were generated, for each boundary condition, the mean performance of the generated geometries was similar or better than that of the test set.

In more detail, in Figure 7, the performance of the test set samples is plotted against the performance of their VAE-reconstructions. The diagrams a, b, and c each correspond to a different boundary condition. In the top row, each geometry instance corresponds to a point on the scatterplot. In the bottom row, the mean and standard deviation of each group of geometries are plotted. The best overall performance would be located in the bottom left corner of the plot. A well-trained VAE should produce reconstructions with performances close to those of the test set. Because the VAE is a probabilistic model, multiple reconstructions were sampled for each instance of the test set (n=100). Additionally, the scatterplot includes the performance of two random geometry generators as baselines for comparison: one uniform random and one Gaussian ($\mu$=5m, $\sigma$=1.5m). Two more baselines are included, coming from simple heuristics: a geometry with a flat horizontal roof and volume equal to the target (optimal volume deviation) and a geometry with a tilted roof at 42° facing south (optimal solar gain).

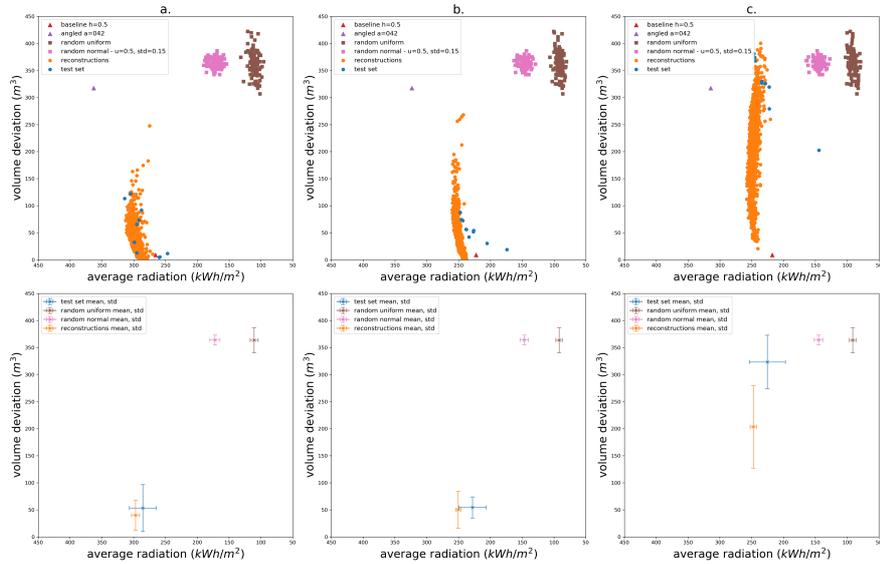

**Fig. 7.** Performance of geometries for three representative cases. Each column a, b, c includes results for a single boundary condition. Geometries from the test set, VAE-reconstructions, baselines, and random generators.

Figure 7a indicates that most reconstructions perform close to the test set or along a curve close to the Pareto front line as implied by the test set samples. The mean performance of the reconstructed geometries is very close to that of the test set geometries. However, in Figure 7b, the distribution of the reconstructions does not follow the one of the test set. The mean solar radiation performance of the reconstructions is better than that of the test set, while the mean volume deviation is approximately the same. Last, in Figure 7c, the performance of the reconstructions is superior in both axes.

## 5.2    Inference Performance

To evaluate the process of navigating the latent space, the performance of the inferred geometries is compared against the performance of the reconstructed geometries. Figure 8 shows representative examples of optimal geometry inference for three different boundary conditions, with varying success. In Figure 8a, the inferred geometries overlap with the reconstructed ones, which means that the optimal geometries as encoded through the VAE were successfully found. For other boundary conditions, such as the one in Figure 8b, the inference is not successful: there is a wide spread of performance for the inferred geometries, with their mean performance located far from that of the reconstructions. Last, in Figure 8c, the performance of many inferred geometries is better than the one of the reconstructions.

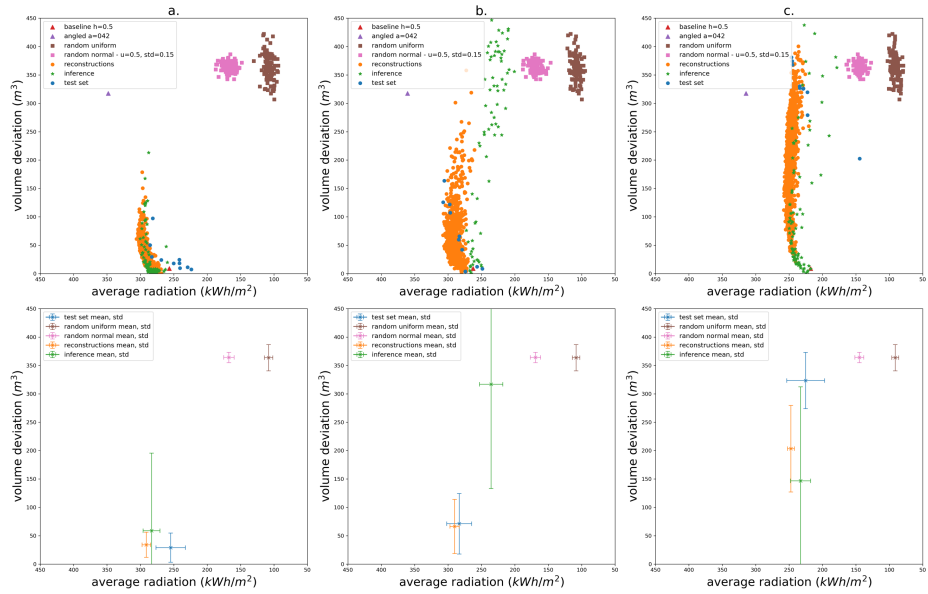

**Fig. 8.** Performance of geometries for three representative cases. Each column a, b, c includes results for a single boundary condition. Geometries from the test set, VAE-inferences, VAE-reconstructions, baselines, and random generators.

The results for all 34 boundary conditions of the test set were coded in the three representative types. Results for 28 boundary conditions were similar to Figure 8a, i.e., successful, 5 were found to be similar to Figure 8b, i.e., not successful, and Figure 8c is the only case of this type.

### 5.3 Hypervolumes

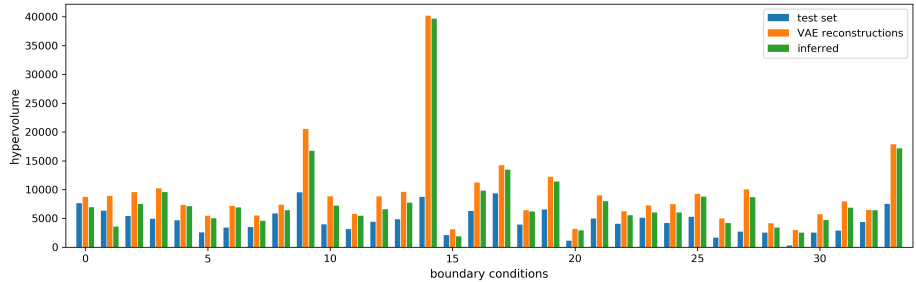

**Fig. 9.** The hypervolumes of the Pareto front of the VAE-reconstructions and the VAE-inferences are plotted against the hypervolume of the test set for each of the 34 unique boundary conditions in the test set.

To further evaluate the generated geometries' performances for both the volume and the solar radiation objectives, the hypervolumes of the Pareto fronts were calculated, as shown in Figure 10. For each boundary condition, three Pareto fronts are compared: the ground truth (test set), the Pareto front of the reconstructed geometries, and the Pareto front of the inferred geometries. The hypervolumes of all three Pareto fronts were calculated using the Python library pymoo [62], using a common reference point for each boundary condition.

The bar graph in Figure 9 shows that for most boundary conditions, the hypervolume of the reconstructions is higher than that of the test set. VAE-generated geometries for boundary condition no. 14 have a significantly higher hypervolume than the test set. This is the same boundary condition as in Figure 7c and Figure 8c. The reason is that the test set geometries for boundary condition 14 are far from optimal. The mean performance of the volume objective is close to random, as the optimization process that generated these geometries got stuck to some local optimum. However, the VAE was able to generalize correctly from higher-quality examples and generated better performing geometries than those found through optimization. These results indicate that the VAE has successfully interpolated the training data, allowing the model to generalize from the training examples to new cases.

Finally, the inferred geometries Pareto front has a hypervolume close to that of the reconstructed ones for most boundary conditions. This confirms that the latent space navigation method, using gradient descent, can successfully find the latent vectors that generate optimally performing geometries.

## 6    Discussion

### 6.1    VAE Sampling for Optimal Solar Gain Performance

The results suggest that a VAE trained on optimally performing geometries can generate geometries with similar performance properties for new simulation contexts. To that extent, the VAE can potentially replace a computationally expensive optimization process, offering a drastic speed improvement. For reference, a single geometry in the dataset typically took approximately 20 minutes to get optimized, while the introduced method used less than 5 seconds.

In the case study, the VAE-generated geometries through the latent space search had a higher hypervolume than those generated through simulated annealing optimization for most boundary conditions. This suggests that, in general, the proposed ML-based method generated higher and more consistent quality results than the optimization method. A potential reason is that during training, as the VAE learns to compress the inputs to a lower-dimensional space, it identifies the essential features to prioritize during the compression. As a result, the higher frequencies – or the noisy information – tend to get lost. On the other hand, a stochastic optimization process such as the SA tends to generate noisy results. Through this process, and by generalizing from all of the training data, the VAE may have filtered out noisy geometric features that were decreasing the solar performances. Finally, the results of the VAE model could be further improved using more extensive and higher quality datasets and hyperparameter tuning.

### 6.2    Beyond Quick Optimization

Apart from quick automatic optimization, the suggested method opens up the potential for optimizing interactively, directly inside the 3d modeling environment. The way that the generative model has been used frees it from any tie to a specific parametric model and any associated limitations. Designer intentions regarding geometric form can be indicated through modeling and used to guide the generation process. For example, a user-designed geometry can easily be encoded as a depth map and guide the latent space search with an appropriate modification of the loss function $L_b$.

### 6.3    Generalizability

The case study demonstrated how a VAE can generate geometries with optimal solar gain and predefined size. However, the suggested method of optimal geometry generation through latent space navigation of a generative model can be used with any performance metric of interest. The optimization workflow described in Section 4.2 could be followed, but individual components such as the geometry representation or the optimization algorithm may be updated to match the needs of each specific problem.

Furthermore, the suggested method is not limited to a single performance objective. The case study already hints at the use of multiple objectives, using the volume target.

### 6.4 Limitations

Concerning the case study, the problem has been simplified and limited in scope in order to facilitate the evaluation of the overall method as well as its individual components. In order to address problems of real-world complexity, a different geometry representation and a more extensive dataset may be needed.

One overall limitation of the suggested method is that the generative model may generate unpredictable results for simulation contexts that are entirely outside of the range of the training set. Appropriate coding of the boundary conditions may alleviate this issue. Additionally, similar to an optimization process, the performance objectives must be specified in advance, i.e., during the training process. The adaptability of the suggested method to changing objectives remains an open question.

## 7 Conclusion

This research introduced a novel method for optimal geometry generation that does not require the designer to use a parametric model. The method aims to provide a more intuitive and interactive alternative for guiding the early phase of performance-driven design than currently available tools. The case study demonstrated the feasibility of using a VAE as a generative model for optimally performing geometries. Future work can focus on expanding the range of the problem variables with real-world problem definition complexity and datasets.

In order to take advantage of the full potential of the suggested method and meet the promise for early-phase design support, future work will also focus on different ways of presenting the results and modes of interactivity inside the modeling environment.

**Acknowledgements**. The first author is partially funded by an Onassis Scholarship (Scholarship ID: F ZO 002/1 – 2018/2019).